\documentclass[10pt, conference, compsocconf]{IEEEtran}
\usepackage{graphicx}
\usepackage{color}
\usepackage{amsmath}
\newcommand{\TODO}[1]{{\color{black}{#1}}}
\ifCLASSINFOpdf
\else
\fi
\begin{document}
%
\title{Emotion Recognition for In-the-wild Videos}

\author{\IEEEauthorblockN{
			Hanyu Liu\IEEEauthorrefmark{1}\IEEEauthorrefmark{3},
			Jiabei Zeng\IEEEauthorrefmark{1} , 
			Shiguang Shan\IEEEauthorrefmark{1}\IEEEauthorrefmark{2} and 
			Xilin Chen\IEEEauthorrefmark{1}\IEEEauthorrefmark{2}
		}
		\IEEEauthorblockA{\IEEEauthorrefmark{1}Key Lab of Intelligent Information Processing, Chinese Academy of Sciences (CAS)\\
			Institute of Computing Technology, CAS, 
			Beijing 100190, China\\ \{jiabei.zeng, sgshan, xlchen\}@ict.ac.cn}
		\IEEEauthorblockA{\IEEEauthorrefmark{2}School of Computer Science and Technology\\
			University of Chinese Academy of Sciences, 
			Beijing 100049, China\\}
		\IEEEauthorblockA{\IEEEauthorrefmark{3}School of Computer Science
		\\Beijing University of Posts and Telecommunications, Beijing 100876, China\\
			\ liuhanyu@bupt.edu.cn}
	}
	

%


\maketitle

\begin{abstract}
This paper is a brief introduction to our submission to the seven basic expression classification track of Affective Behavior Analysis in-the-wild Competition held in conjunction with the IEEE International Conference on Automatic Face and Gesture Recognition (FG) 2020. Our method combines Deep Residual Network (ResNet) and Bidirectional Long Short-Term Memory Network (BLSTM), achieving 64.3\% accuracy and 43.4\% final metric on the validation set.

\end{abstract}


%
\IEEEpeerreviewmaketitle

\section{Introduction}
Automated facial expression recognition (FER) in-the-wild is a long-standing problem in affective computing and human-computer interaction.
To analyze facial expression, psychologists and computer scientists have classified the facial expression into a list of emotion-related categories, such as six basic emotions, i.e., anger, disgust, fear, happiness, sadness, and surprise. 
Ekman et al.\cite{ekman1992argument} have shown that the six basic emotional expressions are universal among human beings.
There has been an encouraging progress on facial expression recognition and during the past decades.

In the the Affective Behavior Analysis in-the-wild (ABAW) 2020 Competition\cite{kollias2020analysing}, the holders provide a large scale in-the-wild database called Aff-Wild2\cite{DBLP:journals/corr/abs-1811-07770, DBLP:journals/corr/abs-1910-04855, DBLP:journals/corr/abs-1811-07771,DBLP:journals/ijcv/KolliasTNPZSKZ19, DBLP:conf/cvpr/KolliasNKZZ17}, including videos annotated with emotion categories, facial action unit\cite{friesen1978facial}, and valence and arousal\cite{russell1980circumplex} dimension. 
In this paper, we present our method used in the expression track in ABAW. 
In this track, the task is to distinguish seven basic facial expressions (i.e., neutral, anger, disgust, fear, happiness, sadness, surprise) of the person in the given videos. 
Our method adopts a 101-layer ResNet\cite{He_2016_CVPR} with convolutional block attention module (CBAM)\cite{woo2018cbam} to extract frame-by-frame features. 
Then, the features are fed into a bidirectional recurrent neural network with long short-term memory (BLSTM)\cite{hochreiter1997long} units. 



\section{Related Work}
Deep neural network-based algorithms are widely used in image and video analysis in recent years. Convolutional neural networks (CNN), for example, residual neural network (ResNet), VGGNet\cite{simonyan2014very}, AlexNet\cite{krizhevsky2012imagenet} are shown effective in image classification and image feature extraction. Long short-term memory network (LSTM)\cite{hochreiter1997long}, a specific improvement of recurrent neural network (RNN), which is capable of capturing serial information, is used in natural language processing as well as video analysis. Meanwhile, the architecture of combination of CNN and RNN is proved to have excellent performance on emotion related tasks. Woo et al. proposed convolutional block attention module (CBAM)\cite{woo2018cbam}, a lightweight and general attention module boosting the performance of all kinds of CNNs. D. Kollias et al. collected Aff-Wild dataset\cite{DBLP:conf/cvpr/ZafeiriouKNPZK17}, the first dataset with annual annotations for each frame of the videos for facial action unit, facial expression and valence-arousal research.

\section{Method}
\begin{figure}[!t]
\centering
\includegraphics[width=3.3in]{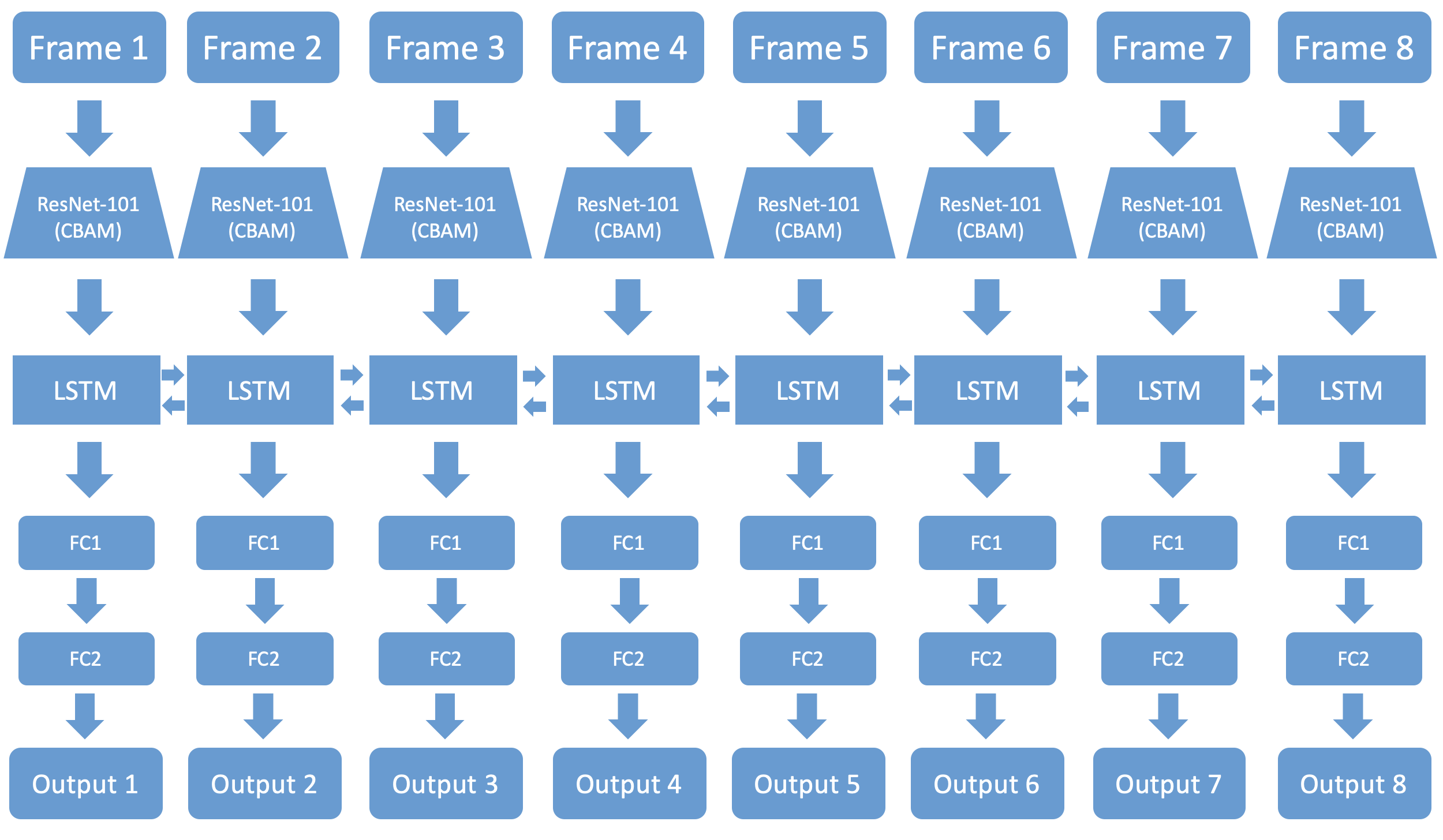}
\caption{Architecture of the proposed method.}
\label{fig:framework}
\end{figure}
Our method consists of three part: ResNet-101 with CBAM that extracts features for each frame, BLSTM that captures the dynamic features of continuous frames, classification module that makes the decisions.
Fig.~\ref{fig:framework} illustrate the framework of our method.
Below, we present the three parts in details. 
\subsection{ResNet-101 with CBAM}
\begin{figure}[!t]
\centering
\includegraphics[width=3.3in]{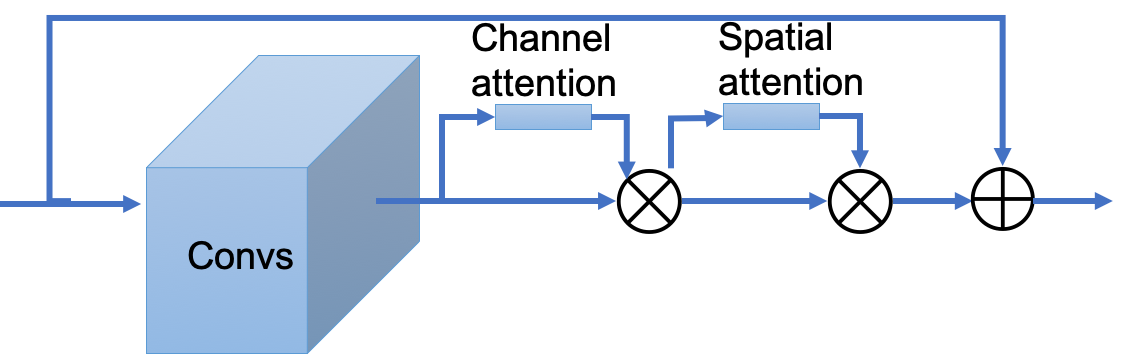}
\caption{Structure of the residual block with CBAM}
\label{fig:block}
\end{figure}

Since ResNet has achieved considerable performance in a lot of computer vision tasks\cite{He_2016_CVPR}, we adopt a 101-layer ResNet (ResNet101) to extract visual features from each frame. 
Considering facial expression appears in particular location in the image, we add a convolutional block attention module (CBAM)\cite{woo2018cbam} after each residual block of ResNet101 to introduce channel attention and spatial attention. Fig.~\ref{fig:block} illustrates the structure of a residual block with CBAM.

\subsection{BLSTM}
Since facial expressions is continuous in the time dimension, we use a Long Short-Term Memory Network (LSTM) to process timing information. Considering that we need to select features for each frame, including the starting frame of the eight frame video clip, we use a bidirectional LSTM here.
\subsection{Classification module}
Lastly, fully connected layers are applied to classify the features into seven classes based on the features extracted and selected by previous layers.

\section{Experiment}
\subsection{Dataset}
Other than the emotion-annotated part in the provided Aff-Wild2 dataset, we used several internal facial expression datasets (AffectNet\cite{mollahosseini2017affectnet}, RAF-DB\cite{li2017reliable}\cite{li2019reliable}) and a self-collected datasets with 300,000 images to pre-train our model. 

{\bf Aff-Wild2:} {Aff-Wild2 annotated in total 539 videos consisting of 2,595,572 frames with 431 subjects, 265 of which are male and 166 female. The dataset are split into train/validation/test parts in a subject-independent manner, with 253, 71, 233 subjects in each.}

{\bf AffectNet: }{AffectNet contains about 440,000 manually annotated facial images collected from Internet search engines. We only used the images with neutral and 6 basic emotions in the training part, including around 280,000 images. }

{\bf RAF-DB:} {Real-world Affective Faces Database (RAF-DB) contains around 30,000 facial images annotated with basic or compound expressions. We only used the 12,271 ones in the training part annotated with basic emotions.}


\subsection{Preprocessing}
The original videos were first divided into frames. These images were later applied on RetinaFace detector\cite{DBLP:journals/corr/abs-1905-00641} to detect all the faced to be analyzed, aligned and cropped into size of $256 \times 256$. 
In order to make the external dataset perform better, all the procedures during preprocessing are similar to the official preprocessing except the tools used. 
For some frames in which human face were not able to be detected by the detector, the corresponding images were removed from the training dataset.

\subsection{Training}
We implemented our model using PyTorch\cite{DBLP:conf/nips/PaszkeGMLBCKLGA19}, on a server with four Nvidia GeForce GTX Titan X GPUs, each with 12GB memory. 
The model is trained with stochastic gradient descent (SGD) with learning rate 0.0001 and momentum 0.9. 
Loss function is cross entropy loss. 
The training batch size is set as 4.
At each step during the  training, one video from all the videos in the training dataset is selected with equal probability. And then a continuous 8 frame video clip (i.e., without any frame from which faces are unable to be detected) is randomly selected from this video as a batch. 
The model makes to its best performance usually within 200,000 batches. 
After every 1000 iteration, we recorded the temporary parameters of the model as a checkpoint.

\subsection{Evaluation}
All the video frames in validation set are arranged into 8 frame clips to be calculated collectively.
If the length of a video is not divisible by 8, the last several frames are padding with zeros. 
The BLSTM part outputs the features for each time step in the clip so that these 8 frames are classified and labeled in a single round. 
We counted the number of successfully predicted frame as well as the total number of frames processed by our model. 

The final metric $\mathcal{S}$ is a combination of accuracy and $F_1$ formulated as:
\begin{align}
\mathcal{S} = 0.33 Acc + 0.67 F_1,
\end{align}
where $Acc$ is the accuracy which is computed as the ration total number of correctly predicted frames over the total frames.
$F_1$ of computed as unweighted mean of all $F_1$ of seven categories. The $F_1$ of a single category is computed as:
\begin{align}
F_1=\frac{2\cdot Precision\cdot Recall}{Precision+Recall}
\end{align}

We manually select the parameter with best performance on validation set from all the checkpoints. 

\subsection{Result}
\begin{table}[!t]
\renewcommand{\arraystretch}{1.3}
\caption{Result on the validation set}
\label{tab:result}
\begin{center}
\begin{tabular}{|c|c|c|c|}
\hline \textbf{Method}&\textbf{Acc}&\textbf{F1}&$\mathcal{S}$\\
\hline baseline\TODO{\cite{kollias2020analysing}}&-&-&0.36\\
 ResNet+BLSTM&0.647&0.281&0.402\\
 ResNet+BLSTM+CBAM&0.640&0.333&0.434\\
\hline 
\end{tabular}
\end{center}
\begin{center}
$\mathcal{S} =0.33Acc + 0.67 F_1$
\end{center}
\end{table}

We evaluated our method on the validation set of Aff-Wild2 and reported the result of our method in Table \ref{tab:result}.
The baseline method is MobileNetV2.
{\bf ResNet+BLSTM} is the combination of vanilla ResNet101 and BLSTM.
{\bf ResNet+CBAM+BLSTM} added the CBAM after each layer of ResNet101.
As can be seen in Table \ref{tab:result}, {\bf ResNet+CBAM+BLSTM} achieves higher final metric $\mathcal{S}$.

\section{Conclusion}
Our proposed method reaches 64.65\% accuracy on the validation set, and 43.43\% final metric on the validation set, 7.43\% higher than the 36\% baseline proposed in the competition announcement.


\section*{Acknowledgment}
The authors would like to thank Xuran Sun for providing us with the pre-trained ResNet FER model and Yuanhang Zhang for assistance.



%
\bibliographystyle{IEEEtran}
\bibliography{bibs}

\end{document}